\documentclass[sigconf, screen]{acmart}

\AtBeginDocument{%
  }

\setcopyright{acmcopyright}
\copyrightyear{2023}
\acmYear{2023}
\acmDOI{XXXXXXX.XXXXXXX}

\acmConference[ACM MM '23]{Make sure to enter the correct
  conference title from your rights confirmation emai}{Oct. 29 – Nov. 3, 2023}{Ottawa, Canada}
\acmPrice{15.00}
\acmISBN{978-1-4503-XXXX-X/18/06}

\renewcommand\footnotetextcopyrightpermission[1]{}

\usepackage{bm}
\usepackage{amsmath}
\usepackage{algorithm}
\usepackage{algorithmic}
\usepackage{multirow}
\usepackage{subcaption}
\usepackage{dsfont}

\begin{document}

\title{Boosting Adversarial Transferability via Fusing Logits of Top-1 Decomposed Feature}

\author{Juanjuan Weng}
\affiliation{%
  \institution{Xiamen University}
  \city{Xiamen}
  \country{China}
}

\author{Zhiming Luo}
\affiliation{%
  \institution{Xiamen University}
  \city{Xiamen}
  \country{China}
}

\author{Dazhen Lin}
\affiliation{%
  \institution{Xiamen University}
  \city{Xiamen}
  \country{China}
}

\author{Shaozi Li}
\affiliation{%
  \institution{Xiamen University}
  \city{Xiamen}
  \country{China}
}

\author{Zhun Zhong}
\affiliation{%
  \institution{University of Trento}
  \city{Trento}
  \country{Italy}
  }



\begin{abstract}
Recent research has shown that Deep Neural Networks (DNNs) are highly vulnerable to adversarial samples, which are highly transferable and can be used to attack other unknown black-box models. To improve the transferability of adversarial samples, several feature-based adversarial attack methods have been proposed to disrupt neuron activation in the middle layers. However, current state-of-the-art feature-based attack methods typically require additional computation costs for estimating the importance of neurons.
To address this challenge, we propose a Singular Value Decomposition (SVD)-based feature-level attack method. Our approach is inspired by the discovery that eigenvectors associated with the larger singular values decomposed from the middle layer features exhibit superior generalization and attention properties. Specifically, we conduct the attack by retaining the decomposed Top-1 singular value-associated feature for computing the output logits, which are then combined with the original logits to optimize adversarial examples.
Our extensive experimental results verify the effectiveness of our proposed method, which can be easily integrated into various baselines to significantly enhance the transferability of adversarial samples for disturbing normally trained CNNs and advanced defense strategies.
The source code of this study is available at \textcolor{blue}{\href{https://github.com/WJJLL/SVD-SSA}{Link}}.
\end{abstract}

\ccsdesc[500]{Security and privacy}
\ccsdesc[500]{Computing methodologies~Image representations}

\keywords{Black-box Attacks; SVD; Adversarial Examples; DNNs}


\maketitle

\section{Introduction}
\begin{figure}[t]
    \centering    \includegraphics[width=0.98\linewidth]{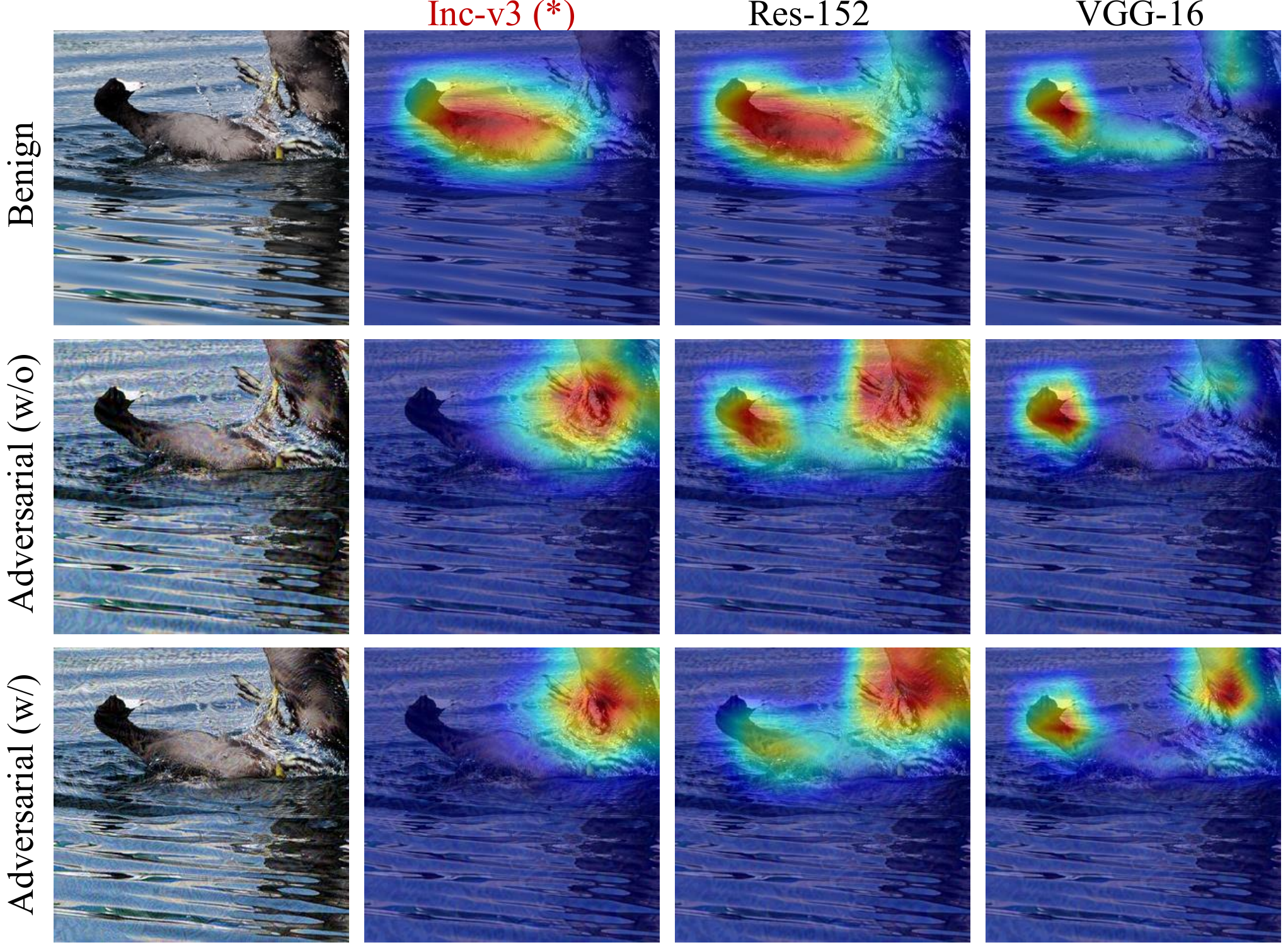}
    \caption{Visualization of attention shift. We apply Eigen-CAM~\cite{muhammad2020eigen} to visualize attention maps of benign (1st row) and adversarial images (2nd and 3rd rows) generated by our method without/with SVD. 
    The 1st row shows maps of three representative models (Inception V3, ResNet-152, and VGG-16) on the benign image. The 2nd and 3rd rows show adversarial images generated on the source model (Inception-V3) and transferred to attack the target models (ResNet-152 and VGG-16). 
    The result demonstrates that adversarial examples with SVD are more capable of disturbing the visual attention from the real foreground object to a common trivial region across different models.
    }
    
    \label{fig:introduction}
\end{figure}

Over the past decade, Deep Neural Networks (DNNs) have achieved significant progress in various computer vision tasks, such as, image classification~\cite{he2016deep}, object detection~\cite{ren2015faster}, and image segmentation~\cite{ronneberger2015u}. However, Goodfellow et al.~\cite{goodfellow2015explaining} have revealed that adding quasi-imperceptible adversarial perturbations into original input images can deceive DNNs into producing incorrect predictions, thus raising severe security issues for many applications, \textit{e.g.}, face recognition~\cite{dong2019efficient} and autonomous driving~\cite{kong2020physgan}. Therefore, it is critical to identify the weakness of the DNNs by developing more effective attack methods beforehand, which can subsequently be used to improve the robustness of DNNs.

Compared to white-box attacks, where the structure and the parameters are known to the attacker, black-box attacks have no access to this information. This makes them more practical for real-world applications. Besides, a more severe issue of adversarial examples is their good transferability~\cite{szegedy2013intriguing}. 
As a result, many works~\cite{dong2018boosting,xie2019improving,wang2022generating} have leveraged the transferability to attack other unknown black-box models. Additionally, there is another type of black-box attack method, \textit{i.e.}, query-based, which conducts several queries to approximate the gradient information for crafting adversarial examples. In this study, we focus on improving the transferability of transfer-based black-box attacks.

Recently, various methods have been developed to improve the transferability of adversarial samples from several different aspects, including gradient-based~\cite{kurakin2018adversarial,lin2020nesterov}, loss function-based~\cite{carlini2017towards,zhao2021success}, input transformation-based~\cite{xie2019improving,byun2022improving}, and feature-based~\cite{huang2019enhancing,wang2021feature,zhang2022improving}. The first three types of methods primarily manipulate either the inputs or outputs of models. In contrast, the last feature-based methods improve the transferability by disrupting the feature maps of intermediate layers in the surrogate model since the important features are typically shared among different models~\cite{wu2020boosting}.

Existing feature-based attacks mainly aim to disturb the neuron activation of adversarial samples compared to original clean samples. The Feature Disruptive Attack (FDA)~\cite{ganeshan2019fda} conducts the attack by optimizing the neuron activation towards mean activation in each middle layer. Instead of treating each neuron equally in the FDA, there are some attempts to consider the importance of each neuron for the attack. The Feature Importance-aware Attack (FIA)~\cite{wang2021feature} estimates the importance by computing the gradient with respect to multiple random masked inputs. The Neuron Attribution-based Attack (NAA)~\cite{zhang2022improving} calculates neuron importance based on the neuron attribute method~\cite{dhamdhere2018important}. Besides, Wu et al.~\cite{wu2020boosting} leverage the attention map to measure the feature importance and regularize optimization by disturbing the attention map. However, these methods often require significant extra computation for obtaining either neuron or feature importance.

On the other hand, a recent study on domain adaptation~\cite{chen2019transferability} analyzed deep features from a spectral perspective and found that the eigenvectors with larger singular values contribute the most to generalization across different domains. Besides, Muhammad et al.~\cite{muhammad2020eigen} proposed the Eigen-CAM for visualizing the attention map of the last convolutional layer. By projecting feature maps on their largest channel-wise eigenvector, the activation map can robustly and reliably localize the foreground objects as shown in Figure~\ref{fig:introduction} (Row 1st). Accordingly, we assume that the eigenvectors corresponding to the largest singular value can be leveraged to increase the adversarial transferability due to its superior generalization and attention properties.

Inspired by the above findings, we propose a new feature-based attack method that utilizes singular value decomposition (SVD). Specifically, our approach involves selecting a middle layer with sufficient semantic information from the surrogate CNN model and decomposing its feature map by the SVD $\bm{X}=\bm{U}\bm{S}\bm{V}^T$. We then only retain the decomposed Top-1 feature with the largest singular value $\bm{Z}=s_1\bm{u}_1\bm{v}_1^T$, and forward it in parallel with the original feature $X$ through all subsequent layers to obtain corresponding output logits, respectively. Finally, these two logits are mixed up to compute the loss function and update adversarial examples via back-propagation. Comparing Row 2 and Row 3 in Figure~\ref{fig:introduction}, we can find that the adversarial sample learned with the SVD has better transferability than the original adversarial sample without SVD. To evaluate the effectiveness of the proposed method, we conduct extensive experiments in both white-box and box-box settings by using different training baselines.

In summary, the main contributions of this study are as follows:
\begin{itemize}
    \item We propose a simple and effective feature decomposition adversarial attack method based on SVD to boost transferability by reliably localizing the critical features in the CNN models. Besides, the proposed feature decomposition attack method can be easily integrated into other attack methods to improve their performance.

    \item Extensive experiments have been conducted to verify the effectiveness of our attack method, which can significantly improve the transferability of adversarial samples against both normally trained CNN models and defense strategies.
\end{itemize}

\section{Related Work}
\subsection{Adversarial Attacks}
In the past few years, many adversarial attack methods~\cite{goodfellow2015explaining,carlini2017towards,kurakin2018adversarial,dong2018boosting,du2020adversarial,mingxing2021towards,wang2022generating,zhang2022improving,byun2022improving} have been proposed for crafting destructive adversarial samples. Besides, adversarial samples exhibit black-box transferability in which adversarial examples from one source model can be reused to fool other unknown models. Thus, it sparks the research of black-box attacks. To generate more destructive and transferable adversarial examples, many methods have been developed that focus on different aspects. We group them into the following categories.

\textbf{Gradient-based:} This type of method increases transferability through advanced gradient-based optimization. The FGSM~\cite{goodfellow2015explaining} is the pioneer gradient-based attack method that directly generates adversarial examples in a single-step optimization by maximizing the cross-entropy loss function. I-FGSM~\cite{kurakin2018adversarial} extends the FGSM into an iterative updating method, which iteratively adds the perturbation with a small step size in the gradient direction. MI-FGSM~\cite{dong2018boosting} incorporates the momentum term into the I-FGSM to encourage a stabilized optimization direction. Similarly, NI-FGSM~\cite{lin2020nesterov} integrates the Nesterov accelerated gradient (NAG) to accumulate the gradients over training iterations. Projected gradient descent (PGD)~\cite{madry2017towards} is a  variant of FGSM, which includes a random initialization within the allowed norm ball and is then followed by the optimization strategy of I-FGSM. LinBP~\cite{guo2020backpropagating} approximates gradients from a smoother surface by skipping over some non-linear activations to reduce the number of non-linear paths in the surrogate model. Huang et al~\cite{huang2022transferable} introduced an integrated gradient to improve transferability, which optimizes the perturbation using a line integral of the gradients from a black reference image to the input image.

\textbf{Loss-based:} Some studies focus on developing new loss functions for optimization to generate adversarial examples.
C\&W loss~\cite{carlini2017towards} directly enlarge the margin distance between the logits of benign examples and the adversarial examples.
In~\cite{naseer2019cross}, a Relativistic Cross-Entropy (RCE) loss is proposed to enforce the confidence between the adversarial sample and the clean sample having a relatively large margin.
\cite{zhao2021success} observed the saturation issue of using the Cross-Entropy loss for crafting targeted adversarial examples and proposes a Logit Loss that is equal to the negative value of target logits for training. For learning disruptive Universal Adversarial Perturbation (UAP), \cite{zhang2021data} leverages the Cosine Similarity Loss for optimization by minimizing the cosine similarity between the real sample and its adversarial counterpart.

\begin{figure*}[t]
    \centering    \includegraphics[width=0.85\linewidth]{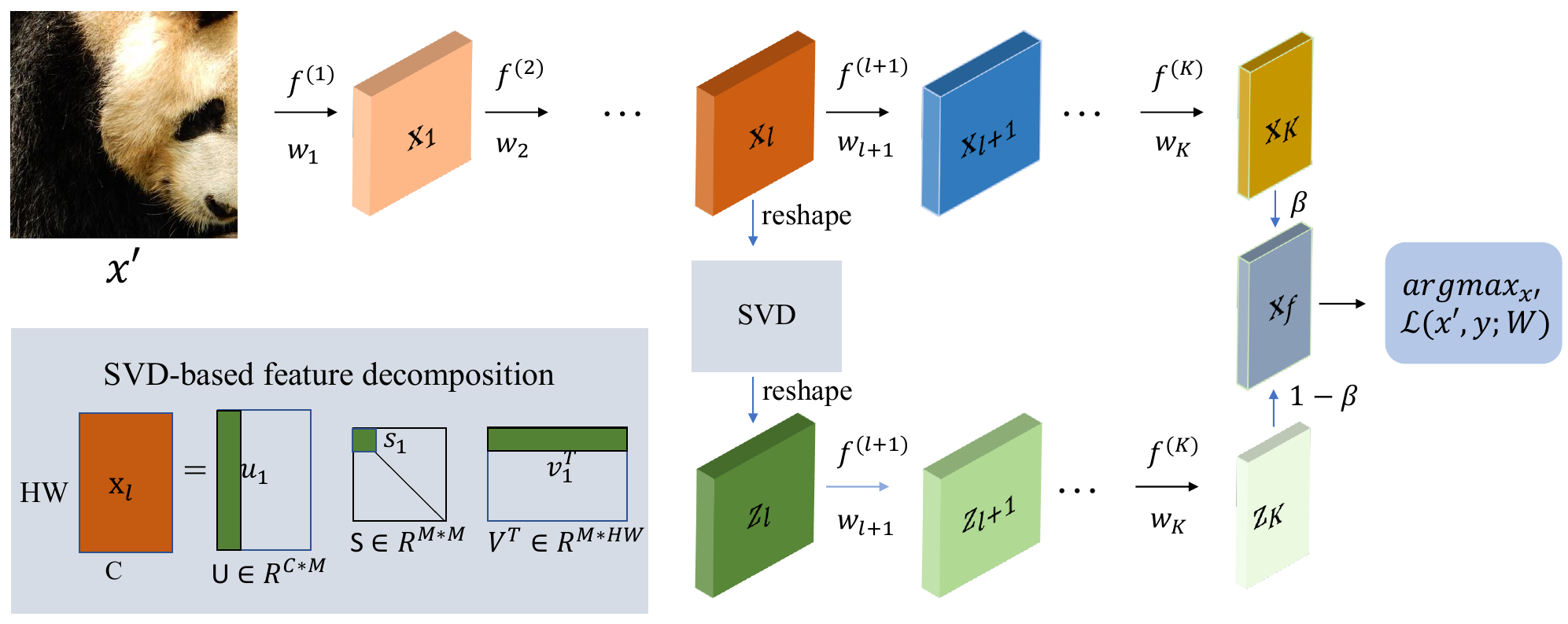}
    \caption{The overall framework of our proposed SVD-based feature decomposition adversarial attack method. The $X_f$ is the final logits, retaining logits $Z_k$ computed by the decomposed Top-1 singular value-related feature $Z_l$ and the original logits $X_k$. }
    \label{fig:framework}
\end{figure*}

\textbf{Input transformation-based:} To further improve the transferability, several input transformation methods have been proposed to augment samples at the input level. For example,  the Diverse Input (DI)~\cite{xie2019improving} applies random resizing and padding to the input images during each training iteration. 
The Translation-Invariant (TI)~\cite{dong2019evading} finds that shifting the input images to obtain an ensemble of adversarial images can improve the performance. To further improve the efficiency, the TI method is implemented by convolving the gradient of the input image with a pre-defined kernel. 
The Scale-Invariant (SI)~\cite{lin2020nesterov} learns the adversarial perturbations over multiple scale copies ($S_i(x)=x/2^i$) of the input image. The Object-Based Diverse Input (ODI)~\cite{byun2022improving} projects the input image onto several 3D objects for data augmentation. Besides, Long et al.~\cite{long2022frequency} proposes a frequency domain data augmentation for training, and this can significantly improve the transferability.
These input transformation-based methods are add-on steps that can be directly integrated into other adversarial attack methods.


\textbf{Feature-based:} Instead of generating adversarial examples mainly based on the final decision of CNN models, some attempts have been made to improve the transferability based on the features of intermediate layers.
Intermediate Level Attack~\cite{huang2019enhancing} enhances transferability using a two-step optimization. The first step involves a standard attack, and the second step increases the perturbation on a pre-specified layer of a CNN model along the optimization direction in the previous step. Feature Disruptive Attack (FDA)~\cite{ganeshan2019fda} generates adversarial examples by disrupting features at each layer of the CNN. Based on the observation that the classification attention maps of different CNN models are very similar, Attention-guided Transfer Attack (ATA)~\cite{wu2020boosting} improves the transferability of adversarial samples by further destructing the attention maps. Feature Importance-aware Attack (FIA)~\cite{wang2021feature} uses multiple random masked input images to compute feature importance based on the accumulated gradients w.r.t an intermediate layer, and then compute the perturbation by disrupting the weighted feature map. Neuron Attribution-based Attack (NAA)~\cite{zhang2022improving} computes neuron importance in a middle layer based on neuron attributes and conducts the feature-level attacks by weighting neurons.

Compared with other feature-based methods, our method does not actually require explicit computation of the feature importance of different neurons. Instead, the SVD decomposition can implicitly identify important features that improve transferability. Additionally, our proposed approach is an add-on component that can be incorporated into various different attacking methods to enhance their performance.

\subsection{Adversarial Defense}
Recently, numerous methods have been proposed to mitigate the security issue raised by adversarial attacks. These methods can be mainly divided into adversarial training-based and input pre-processing methods. 
The adversarial training-based method~\cite{madry2017towards} is a simple and effective way of improving the robustness of the model, where the adversarial samples are augmented with the clean training samples to retrain the model. On the other hand, the input pre-processing methods mainly aim to cure the infection of adversarial perturbations before feeding the images into CNNs, \textit{e.g.}, JPEG compression~\cite{guo2017countering}, random resizing and padding (R\&P)~\cite{xie2017mitigating}, and 
high-level representation guided denoiser (HGD)~\cite{lin2020nesterov}. Additionally, to further improve defense performance, many works try to combine the benefits of adversarial training and input pre-processing methods, \textit{e.g.}, NeurIPS-r3 solution~\cite{thomas2017defense} and Neural Representation
Purifier (NRP)~\cite{naseer2020self}. In this study, we leverage these defenses to validate the disruptiveness of our proposed adversarial attack.

\section{Preliminaries}

We consider Convolutional Neural Networks (CNNs) used for image classification as a function $f$, which can be described as a series of successive functions $f = f^{(K)} \circ f^{(K-1)} \circ \cdots  \circ f^{(1)}$ with parameters $W = (\textbf{w}_K, \textbf{w}_{k-1}, \cdots, \textbf{w}_1 )$. Here, $f^{(l)}$ is layer $l$, and $\textbf{w}_l$ are its corresponding parameters. $K$ represents the total number of layers.  Besides, we denote the output of layer $l$ as ${\textbf{X}}_{l}$ for an input image $x$.

Given a CNN model $f$ and a clean image $x$ with ground truth label $y$, our primary goal is to craft an adversarial example $x'$ that can mislead the classifier $f$ into making an unexpected prediction. To ensure that the induced perturbation to the original image $x$ is minimally changed, an adversarial example is usually bounded in the $l_p$-norm ball centered at $x$ with radius $\epsilon$. Following previous works~\cite{goodfellow2015explaining,xie2019improving,gao2021feature}, we adopt the widely advocated $l_\infty$-norm in this work. More formally, it can be summarized as follows: 
\begin{equation}
    \arg\max f(x') \neq y ,  \quad 
 \mbox{s.t.} \quad ||x'-x||_{\infty} \leq \epsilon.
 \label{eq:def}
\end{equation}
Let $\mathcal{L}(x',y;W)$ represent the loss function for crafting adversarial examples in Eq.~\ref{eq:def}, which can be formulated as:
\begin{equation}
    \arg\max_{x'}  \mathcal{L}(x',y;W),
 \label{eq:def_loss}
\end{equation}
where $\mathcal{L}(x',y;W)$ is usually the cross-entropy loss. 

In real-world scenarios, the attacker has no access to the parameter $W$ of the victim model $f$. Following the common practice, we fool the victim model based on the transferability of the adversarial examples, crafted by a substitute model. Besides, we enhance transferability based on advanced gradient-based optimization methods, \textit{e.g.}, Iterative-Fast Gradient Sign Method (I-FGSM). The I-FGSM iteratively adds the perturbation with a small step size $\alpha$ in the gradient direction:
\begin{equation}
   {{x'}}_{0} = x, \quad {{x'}}_{i+1} = {{x}}'_{i} + \alpha \cdot \text{sign}(\nabla_{{x'}}  \mathcal{L}(x'_{i},y;W)),
\end{equation}
where ${x}_{i}'$ denotes adversarial image in the $i_{th}$ iteration, and $\alpha=\epsilon/T$ ensures that $x'$ is within the $l_p$-norm when optimized by a maximum of $T$ iterations.

\section{Method}

In this section, we introduce our proposed attack method based on the SVD. We extract the Top-1 singular value feature from a middle layer $l$ to obtain more generalized information for boosting the transferability of adversarial examples. Finally, we describe the loss function for optimization.

\subsection{Overall architecture}
The overall architecture of our proposed feature decomposition attack framework is shown in Fig.~\ref{fig:framework}. For an input image $x$, we pass it through the surrogate CNN model to compute the feature of each layer $f^{(i)}$. In a middle layer $l$, we decompose the corresponding feature $\bm{X}_l$ by SVD and obtain the Top-1 feature $\bm{Z}_l$ with the largest singular value. Then, the original feature $\bm{X}_l$ and the top-1 feature $\bm{Z}_l$ are concurrently transmitted to the remaining layers of CNN to compute the corresponding outputs $\bm{X}_k$ and $\bm{Z}_k$ in two parallel branches, respectively. Finally, we compute the attack loss function based on $\bm{X}_k$ and $\bm{Z}_k$, and update the adversarial example by gradient backpropagation.

\subsection{SVD-based feature decomposition}

A recent study~\cite{chen2019transferability} analyzed the deep feature from a spectral analysis perspective for domain adaptation tasks. After decomposing the feature representation by SVD, the eigenvectors associated with larger singular values usually contribute to generalization in cross-domain classification. On another aspect, \cite{muhammad2020eigen} proposed an Eigen-CAM for visualizing the learned features in a CNN. By projecting the feature map of the last convolutional layer on its largest channel-wise eigenvector, the activation map can robustly and reliably localize the foreground objects without the computation of back-propagation. 

To enhance the transferability of attacking unknown black-box CNNs, it is encouraged to address the overfitting issue of adversarial samples biasing towards the training surrogate CNN. Inspired by findings in~\cite{chen2019transferability,muhammad2020eigen}, we also utilize the feature decomposition based on SVD to enhance the adversarial transferability. The detailed computation procedures are as follows.

Given the feature map $\bm{X}_l\in R^{C \times H \times W}$ of middle layer $l$, we first reshape it to the size $\mathcal{X}_l\in R^{C \times HW}$. Let $M= \min(C, HW)$, we then perform the Singular Value Decomposition (SVD) to decompose the feature into:
\begin{equation}
    \mathcal{X}_l = \bm{U}\bm{S}\bm{V}^T,
\end{equation}
where $\bm{S}\in R^{M\times M}$ is the rectangular diagonal singular value matrix, $\bm{U}\in R^{C\times M}$ is the left orthogonal singular vector matrix, $\bm{V}\in R^{HW\times M}$ is the right orthogonal singular vector matrix.

After the decomposition, we compute the Top-1 decomposed feature $\bm{Z}_l$ with respected to the largest singular value $s_1$ by:
\begin{equation}
    \bm{Z}_l = s_1 \bm{u}_1 \bm{v}_1^T,
\end{equation}
where $\bm{u}_1$ and  $\bm{v}_1$ are the corresponding left and right singular vectors, respectively. The $\bm{Z}_l$ is reshaped back to $\bm{Z}_l\in R^{C \times H \times W}$ and is  further used for computing the decomposed logits. Besides, we evaluate the attacking performance of using different Top-$k$ singular value to obtain the decomposed feature, \textit{i.e.}, $\bm{Z}_l = \sum_{i=1}^k s_i \bm{u}_i \bm{v}_i^T$.




\subsection{Attacking algorithm}
Since our proposed SVD-based feature decomposition can be integrated into different adversarial baselines for crafting the adversarial samples, this section only presents the loss function used for learning the adversarial perturbation in this study. 

After getting the Top-1 decomposed feature $\bm{Z}_l$, we jointly forward it with $\bm{X}_l$ in parallel through the rest of the layers in the CNNs, and we can obtain the output logits $\bm{Z}_k$ and $\bm{X}_K$, respectively. Then, we combine them jointly to obtain the final logits:
\begin{equation}
    \bm{X}_f = \beta \bm{X}_K + (1-\beta) \bm{Z}_k,
    \label{eq:combine_logits}
\end{equation}
where $\beta$ is a hype-parameter controlling the influence of the decomposed feature.

Based on the fused logits $\bm{X}_f$, we can compute the final loss function by the cross-entropy:
\begin{equation}
    \mathcal{L}(x',y;W) = -\mathds{1}_{y}^T \cdot \log \big( \text{softmax}(\bm{X}_f) \big),
\end{equation}
where $y$ is the ground-truth label of the input $x$.

\section{Why SVD method is effective?}
In this section, we will analyze why incorporating SVD to craft adversarial examples can improve attacking performance. Representation similarity (RS) metrics have been widely used to quantitatively compare the activations of different sets of neurons on a given dataset. Following~\cite{cianfarani2022understanding}, we leverage the linear Centred Kernel Alignment (CKA) to measure the RS between the clean samples and adversarial samples. The linear CKA is computed as follows:
\begin{equation}
    \text{CKA}_{L} (\bm{X}, \bm{Y}) = \frac{||\bm{Y}^T\bm{X}||^2_F}{||\bm{X}^T\bm{X}||_F||\bm{Y}^T\bm{Y}||_F},
\end{equation}
where $\bm{X}\in \mathbb{R}^{n\times d_1}$ and $\bm{Y}\in \mathbb{R}^{n\times d_2}$ are two feature activations of a set with $n$ samples. The $d_1$ and $d_2$ are the corresponding feature dimensions. To compute the linear CKA, we craft adversarial examples for 1000 images from ImageNet-Compatible Dataset by using Inc-v3 as the source model. The S$^{2}$I-TI-DIM~\cite{long2022frequency} is used for optimization. More training details can be found in the experiments.

First, we compute the linear CKA between the features of clean samples and adversarial samples from the source Inc-v3 model. At layer $l$, we denote the features of all clean samples, adversarial samples (w/o SVD), and adversarial samples (w/ SVD) as $\bm{X} \in \mathbb{R}^{1000\times d_l}$, $\Tilde{\bm{X}} \in \mathbb{R}^{1000\times d_l}$, $\Tilde{\bm{X}}_{svd} \in \mathbb{R}^{1000\times d_l}$, where $d_l$ is the feature dimension at layer $l$. Then, we compute $\text{CKA}_{L} (\bm{X}, \Tilde{\bm{X}})$ and $\text{CKA}_{L} (\bm{X}, \Tilde{\bm{X}}_{svd})$.
As shown in Table~\ref{tab:cka_clean}, we can find that   the similarities between the representations of clean and adversarial samples (w/ SVD) are much smaller than without SVD, particularly at those higher layers. These results indicate that the adversarial examples crafted by using SVD are more disruptive in terms of attacking the white-box model.

Additionally, according to ~\cite{cianfarani2022understanding}, "high transferability indicates that the target network has learned similar decision boundary". Therefore, when the representations that are correlated with the classification are more similar, the adversarial examples trained on the source models are easier to transfer to the target models. Specifically, we compute the linear CKA at the final global pooling layer and FC layer. These layers are highly correlated with the classification since different models have different architectures.  In this part, we directly use the adversarial samples generated by Inc-v3 to compute the features from Inc-v3, ResNet-152, and VGG-16. As shown in Table~\ref{tab:cka_model}, the linear CKA between Inc-v3 and ResNet-152 at Pool and FC layers for 1000 clean samples are 0.3769 and 0.4407, respectively. The linear CKA of adversarial samples (w/o SVD) will decrease to 0.2066 and 0.2182. On the other hand, we can find that the linear CKA of adversarial samples (w/ SVD) will slightly increase to 0.2273 and 0.2447, which means the representation of adversarial samples (w/ SVD) transferred from Inc-v3 to ResNet-152 are more similar than adversarial samples (w/o SVD). Besides, a similar finding can be found between Inc-v3 and VGG-16. These results indicate that our proposed SVD-based optimation methods can enhance the classification-related representation similarity of the adversarial examples between the source model and the target model, thereby improving transferability.

Upon analyzing the representation similarity using linear CKA, we can observe that crafting adversarial examples using SVD is effective in disrupting feature activations in the white-box model. Additionally, this also can enhance the transferability of adversarial samples for attacking black-box models.

\begin{table}[t]
\caption{The linear CKA between the features of clean samples and adversarial samples computed from different layers of the source Inc-v3 model.}
\label{tab:cka_clean}
\resizebox{\columnwidth}{!}{
\begin{tabular}{c|cccccc}
\hline
         & Conv2d\_4a & Mixed\_5d & Mixed\_6e & Mixed\_7c & Pool   & FC     \\ \hline
w/o      & 0.9279          & 0.7183    & 0.6138    & 0.2765    & 0.1278 & 0.0885 \\
w/       & 0.9251          & 0.6902    & 0.4697    & 0.2120    & 0.1140 & 0.0731 \\\hline
\end{tabular}
}
\end{table}

\begin{table}[t]
\caption{The linear CKA between the features under the setting of transferring the adversarial examples crafted by Inc-v3 to ResNet-152 and VGG-16.}
\label{tab:cka_model}
\resizebox{\columnwidth}{!}{
\begin{tabular}{c|ccc|ccc}
\hline
     & \multicolumn{3}{c|}{Inc-v3 $\rightarrow$ ResNet-152}       & \multicolumn{3}{c}{Inc-v3 $\rightarrow$ VGG-16}                              \\
     & Clean  & Adv~(w/o) & Adv~(w/)                    & Clean                      & Adv~(w/o)                   & Adv~(w/) \\ \hline
Pool & 0.3769 & 0.2066   & 0.2273 & 0.2160 & 0.1107 & 0.1165 \\
FC   & 0.4407 & 0.2182   & 0.2447 & 0.3196 & 0.1793 & 0.1896 \\ \hline 
\end{tabular}
}
\end{table}

\section{Experiments}
\subsection{Experimental Setup}

\textbf{Dataset:} 
Following previous studies~\cite{dong2018boosting,long2022frequency}, we conduct the experiments on the challenge ImageNet-Compatible Dataset\footnote{\url{https://github.com/cleverhans-lab/cleverhans/tree/master/cleverhans_v3.1.0/examples/nips17_adversarial_competition/dataset}}. The dataset contains 1,000 images with 1,000 unique class labels that correspond to the original ImageNet dataset. 

\textbf{Models:} In this study, we utilize four normally trained ImageNet CNN models as source models to craft adversarial examples, \textit{i.e.}, Inception-v3 (Inc-v3)~\cite{szegedy2016rethinking}, Inception-v4 (Inc-v4)~\cite{szegedy2017inception}, Inception-Resnet-v2 (IncRes-v2)~\cite{szegedy2017inception} and Resnet-v2-152 (Res-152)~\cite{he2016deep}. To evaluate the performance of the adversarial samples, we use the four training source models and Resnet-v2-50 (Res-50) and Resnet-v2-101 (Res-101) as victim models. Besides, for further evaluation of attacking defense models, we select four adversarially trained models based on PGD-AT~\cite{madry2017towards} and Ensemble Adversarial Training~\cite{tramer2017ensemble}, which augments a model’s training data with adversarial examples crafted on other static pre-trained models for yielding models with stronger robustness to black-box attacks. Specifically, adversarially trained models include an adversarially trained Inception-v3 model (Inc-v3$_{adv}$), an ensemble of three adversarially trained Inception-v3 models (Inc-v3$_{ens3}$), an ensemble of four adversarially trained Inception-v3 models (Inc-v3$_{ens4}$) and an ensemble of three adversarially trained Inception-Resnet-v2 models (IncRes-v2$_{ens}$). Additionally, we select advanced defense methods that cover random resizing and padding (R\&P)~\cite{xie2017mitigating}, NIPS-r3~\cite{thomas2017defense}, HGD~\cite{lin2020nesterov},  JPEG~\cite{guo2017countering}, randomized smoothing (RS)~\cite{jia2019certified} and NRP~\cite{naseer2020self} ), which are robust against black-box attacks.

\textbf{Baseline Method:} To thoroughly evaluate performance, we integrate our method into various well-known adversarial attacking baselines, including DI-FGSM~\cite{xie2019improving}, TI-FGSM~\cite{dong2019evading}, MI-FGSM~\cite{dong2018boosting}, VT-FGSM~\cite{wang2021enhancing}, SI-NI-FGSM~\cite{lin2020nesterov}, PI-TI-DI-FGSM~\cite{gao2020patch}, and Admix~\cite{wang2021admix}. Additionally, we incorporate our method into one current SOAT attacking method S$^{2}$-I-FGSM~\cite{long2022frequency}, which adopts a frequency-based spectral transformation for data augmentation. Besides, we combine these simple baselines jointly to construct stronger disruptive attacking baselines.

\begin{table*}[t]
 \caption{The attack success rates ($\%$) on six normally trained models. The adversarial examples are crafted via Inc-v3, Inc-v4, IncRes-v2, and Res-152, respectively.}
 \label{tab:Normally-Trained}
\begin{tabular}{c|c|c|cccccc|c}
\hline \hline
Model & Attack & SVD & Inc-v3 & Inc-v4 & IncRes-v2 & Res-152 & Res-50 & Res-101 & Avg \\
\hline
\multirow{10}{*}{Inc-v3} & MI-FGSM & w/o & 100.0 & 50.6 & 45.8 & 41.2 & 47.1 & 42.1 & 54.5 \\
 &  & w/ & 100.0 & 55.2 & 52.8 & 44.9 & 49.9 & 46.5 & 58.2 \textbf{($\uparrow 3.7$}) \\
  \cline{2-10}
 & DI-FGSM & w/o & 99.9 & 49.6 & 37.7 & 32.7 & 37.4 & 33.5 & 48.5 \\
 &  & w/ & 99.5 & 53.1 & 44.7 & 34.2 & 43.0 & 37.6 & 52.0 \textbf{($\uparrow 3.5$)}\\
    \cline{2-10}
 & S2I-FGSM & w/o & 99.7 & 63.6 & 57.5 & 49.9 & 56.5 & 52.0 & 63.2 \\
 &  & w/ & 99.8 & 69.8 & 65.9 & 57.2 & 64.2 & 60.1 & 69.5 \textbf{($\uparrow 6.3$)}\\
    \cline{2-10}
 & SI-NI-FGSM & w/o & 100.0 & 76.9 & 74.7 & 68.0 & 72.2 & 69.1 & 76.8 \\
 &  & w/ & 100.0 & 80.2 & 79.1 & 70.7 & 76.2 & 72.3 & 79.8 \textbf{($\uparrow 3.0$)}\\
    \cline{2-10}
 & VT-MI-FGSM & w/o & 100.0 & 75.0 & 69.9 & 62.6 & 66.8 & 62.3 & 72.8 \\
 &  & w/ & 99.8 & 80.6 & 78.7 & 69.0 & 73.7 & 70.6 & 78.7 \textbf{($\uparrow 5.9$)}\\
 \hline
\multirow{10}{*}{Inc-v4} & MI-FGSM & w/o & 60.7 & 99.9 & 45.5 & 42.3 & 47.7 & 43.5 & 56.6 \\
 &  & w/ & 63.5 & 100.0 & 48.7 & 44.9 & 50.1 & 46.3 & 58.9 \textbf{($\uparrow 2.3$)}\\
    \cline{2-10}
 & DI-FGSM & w/o & 52.5 & 99.2 & 36.0 & 29.9 & 32.7 & 30.5 & 46.8 \\
 &  & w/ & 56.0 & 99.5 & 40.2 & 32.6 & 39.2 & 33.7 & 50.2 \textbf{($\uparrow 3.4$)}\\
    \cline{2-10}
 & S2I-FGSM & w/o & 71.5 & 99.6 & 55.4 & 49.0 & 57.1 & 48.9 & 63.6 \\
 &  & w/ & 76.1 & 99.6 & 62.6 & 54.3 & 60.4 & 55.3 & 68.1 \textbf{($\uparrow 4.5$)}\\
    \cline{2-10}
 & SI-NI-FGSM & w/o & 86.0 & 100.0 & 79.1 & 73.6 & 77.1 & 73.7 & 81.6 \\
 &  & w/ & 87.1 & 100.0 & 81.4 & 75.3 & 79.6 & 75.9 & 83.2 \textbf{($\uparrow 1.6$)}\\
   \cline{2-10}
 & VT-MI-FGSM & w/o & 80.1 & 99.8 & 70.6 & 64.2 & 65.8 & 63.8 & 74.1 \\
 &  & w/ & 85.0 & 99.9 & 76.5 & 69.9 & 71.4 & 68.6 & 78.6 \textbf{($\uparrow 4.5$)}\\
 \hline
\multirow{10}{*}{IncRes-v2} & MI-FGSM & w/o & 60.2 & 53.4 & 99.2 & 44.1 & 49.6 & 46.7 & 58.9 \\
 &  & w/ & 71.5 & 61.2 & 99.7 & 50.8 & 57.4 & 54.8 & 65.9 \textbf{($\uparrow 7.0$)}\\
    \cline{2-10}
 & DI-FGSM & w/o & 56.3 & 51.6 & 97.7 & 37.1 & 39.8 & 36.3 & 53.1 \\
 &  & w/ & 64.2 & 58.8 & 98.9 & 39.6 & 46.0 & 41.9 & 58.2 \textbf{($\uparrow 5.1$)}\\
    \cline{2-10}
 & S2I-FGSM & w/o & 77.3 & 68.0 & 98.4 & 56.1 & 62.4 & 58.1 & 70.1 \\
 &  & w/ & 80.0 & 72.3 & 99.1 & 60.9 & 66.6 & 62.6 & 73.6 \textbf{($\uparrow 3.5$)}\\
   \cline{2-10}
 & SI-NI-FGSM & w/o & 86.6 & 84.0 & 99.8 & 76.5 & 80.0 & 77.5 & 84.1 \\
 &  & w/ & 90.0 & 87.8 & 100.0 & 81.8 & 84.6 & 83.6 & 88.0 \textbf{($\uparrow 3.9$)}\\
    \cline{2-10}
 & VT-MI-FGSM & w/o & 81.1 & 76.6 & 99.0 & 66.7 & 70.1 & 68.7 & 77.0 \\
 &  & w/ & 88.0 & 84.9 & 99.7 & 77.1 & 79.2 & 78.4 & 84.6 \textbf{($\uparrow 7.6$)}\\
 \hline
\multirow{10}{*}{Res-152} & MI-FGSM & w/o & 56.3 & 50.8 & 46.0 & 99.4 & 82.9 & 86.2 & 70.3 \\
 &  & w/ & 63.9 & 53.9 & 50.1 & 99.4 & 88.0 & 89.1 & 74.1 \textbf{($\uparrow 3.8$)}\\
  \cline{2-10}
 & DI-FGSM & w/o & 55.9 & 53.2 & 46.3 & 99.6 & 83.2 & 84.3 & 70.4 \\
 &  & w/ & 61.4 & 56.9 & 50.4 & 99.5 & 86.3 & 88.5 & 73.8 \textbf{($\uparrow 3.4$)}\\
   \cline{2-10}
 & S2I-FGSM & w/o & 66.6 & 63.3 & 57.6 & 99.8 & 93.5 & 93.9 & 79.1 \\
 &  & w/ & 71.6 & 64.9 & 62.5 & 99.8 & 95.2 & 95.2 & 81.5 \textbf{($\uparrow 2.4$)}\\
  \cline{2-10}
 & SI-NI-FGSM & w/o & 76.3 & 72.5 & 72.6 & 99.8 & 94.2 & 95.3 & 85.1 \\
 &  & w/ & 78.5 & 76.5 & 75.9 & 99.9 & 95.5 & 95.9 & 87.0 \textbf{($\uparrow 1.9$)}\\
  \cline{2-10}
 & VT-MI-FGSM & w/o & 73.7 & 68.7 & 65.8 & 99.5 & 93.2 & 93.8 & 82.5 \\
 &  & w/ & 80.2 & 75.9 & 73.4 & 99.5 & 93.9 & 95.7 & 86.4 \textbf{($\uparrow 3.9$)}\\
 \hline \hline
\end{tabular}
\end{table*}

\begin{table*}[t]
\centering
\caption{The attack success rates (\%) on ten defenses. The adversarial examples
are crafted via Inc-v3, Inc-v4, IncRes-v2 and Res-152.}
\label{tab:defense}
\resizebox{\linewidth}{!}{
\begin{tabular}{c|c|c|cccccccccc|c}
 \hline\hline
Model & Attack & SVD &Inc-v3$_{adv}$ & Inc-v3$_{ens3}$ & Inc-v3$_{ens4}$ & IncRes-v2$_{ens}$ & HGD & R\&P & NIPS-r3 & JPEG & RS & NRP & AVG. \\
\hline
\multirow{12}{*}{Inc-v3} & \multirow{2}{*}{TI-DIM} & w/o & 45.5 & 43.1 & 43.2 & 28.6 & 36.2 & 30.0 & 35.4 & 54.6 & 36.2 & 24.9 & 37.8 \\
 &  & w/ & 49.2 & 48.4 & 45.5 & 32.0 & 40.0 & 34.3 & 41.2 & 59.8 & 39.0 & 24.3 & 41.4 \\
 \cline{2-14}
 & \multirow{2}{*}{PI-TI-DI-FGSM} & w/o & 44.6 & 44.0 & 46.0 & 33.4 & 32.4 & 33.7 & 37.0 & 46.7 & 64.7 & 36.9 & 41.9 \\
 &  & w/ & 48.0 & 46.6 & 47.9 & 35.8 & 34.2 & 36.5 & 40.3 & 50.2 & 66.7 & 38.9 & 44.5 \\
  \cline{2-14}
& \multirow{2}{*}{VT-TI-DIM} & w/o & 64.3 & 65.0 & 64.3 & 47.6 & 58.8 & 51.7 & 57.6 & 74.9 & 42.6 & 35.2 & 56.2 \\
 &  & w/ & 67.9 & 69.4 & 67.3 & 51.9 & 61.7 & 53.5 & 62.3 & 80.2 & 45.1 & 37.8 & 59.7 \\
  \cline{2-14}
& \multirow{2}{*}{Admix-TI-DIM} & w/o & 77.0 & 75.6 & 73.9 & 60.0 & 69.6 & 62.2 & 70.3 & 84.7 & 55.9 & 46.3 & 67.6 \\
 &  & w/ & 81.6 & 82.7 & 79.3 & 64.8 & 73.4 & 57.5 & 75.5 & 88.2 & 58.9 & 47.3 & 70.9 \\
  \cline{2-14}
& \multirow{2}{*}{S$^{2}$I-TI-DIM} & w/o & 82.8 & 81.9 & 79.2 & 68.8 & 77.1 & 70.0 & 75.9 & 87.2 & 61.2 & 57.2 & 74.1 \\
 &  & w/ & 85.8 & 84.8 & 83.5 & 72.9 & 80.6 & 75.1 & 80.1 & 88.9 & 64.4 & 59.1 & 77.5 \\
 \hline
\multirow{12}{*}{Inc-v4} & \multirow{2}{*}{TI-DIM} & w/o & 38.8 & 38.9 & 37.1 & 29.1 & 36.3 & 29.9 & 33.4 & 48.7 & 35.9 & 19.2 & 34.7 \\
 &  & w/ & 41.3 & 42.3 & 41.6 & 29.7 & 37.0 & 30.8 & 36.2 & 53.3 & 37.5 & 21.3 & 37.1 \\
   \cline{2-14}
 & \multirow{2}{*}{PI-TI-DI-FGSM} & w/o & 40.6 & 41.6 & 44.2 & 34.4 & 31.7 & 33.1 & 32.0 & 45.4 & 63.5 & 34.2 & 40.1 \\
 &  & w/ & 43.6 & 44.8 & 45.8 & 35.2 & 33.4 & 35.6 & 37.8 & 47.7 & 65.5 & 36.9 & 42.6 \\
   \cline{2-14}
 & \multirow{2}{*}{VT-TI-DIM} & w/o & 57.9 & 61.4 & 59.7 & 48.4 & 58.8 & 51.9 & 55.1 & 70.8 & 41.3 & 32.0 & 53.7 \\
 &  & w/ & 61.3 & 64.7 & 62.7 & 52.8 & 61.8 & 53.6 & 59.1 & 76.1 & 42.0 & 36.1 & 57.0 \\
   \cline{2-14}
 & \multirow{2}{*}{Admix-TI-DIM} & w/o & 77.3 & 78.7 & 74.5 & 65.7 & 75.8 & 69.2 & 74.0 & 83.7 & 54.5 & 49.1 & 70.3 \\
 &  & w/ & 78.7 & 81.2 & 77.0 & 69.0 & 76.7 & 72.3 & 75.8 & 86.2 & 56.0 & 47.6 & 72.1 \\
   \cline{2-14}
 & \multirow{2}{*}{S$^{2}$I-TI-DIM} & w/o & 80.2 & 80.8 & 77.7 & 71.0 & 77.2 & 72.4 & 76.4 & 84.7 & 60.4 & 56.5 & 73.7 \\
 &  & w/ & 82.1 & 82.2 & 78.8 & 72.9 & 78.3 & 74.3 & 78.2 & 86.3 & 63.6 & 59.0 & 75.6 \\
 \hline
\multirow{12}{*}{IncRes-v2} & \multirow{2}{*}{TI-DIM} & w/o & 43.1 & 46.4 & 40.9 & 39.4 & 43.4 & 39.0 & 41.5 & 56.6 & 36.7 & 22.2 & 40.9 \\
 &  & w/ & 52.3 & 56.8 & 50.2 & 41.6 & 50.2 & 45.2 & 50.9 & 68.6 & 40.4 & 26.9 & 48.3 \\    \cline{2-14}
 & \multirow{2}{*}{PI-TI-DI-FGSM} & w/o & 47.8 & 48.7 & 49.5 & 45.6 & 39.4 & 45.8 & 46.5 & 51.1 & 66.3 & 40.9 & 48.2 \\
 &  & w/ & 54.9 & 54.7 & 56.1 & 50.3 & 45.7 & 50.2 & 51.6 & 56.1 & 70.3 & 45.8 & 53.6 \\    \cline{2-14}
 & \multirow{2}{*}{VT-TI-DIM} & w/o & 61.8 & 66.4 & 61.7 & 59.5 & 64.0 & 61.4 & 62.1 & 72.6 & 43.4 & 35.4 & 58.8 \\
 &  & w/ & 72.8 & 78.0 & 71.7 & 67.8 & 72.4 & 69.7 & 73.1 & 84.7 & 47.4 & 42.8 & 68.0 \\    \cline{2-14}
 & \multirow{2}{*}{Admix-TI-DIM} & w/o & 83.5 & 84.8 & 83.4 & 79.0 & 82.9 & 81.6 & 82.3 & 88.5 & 64.0 & 61.2 & 79.1 \\
 &  & w/ & 88.0 & 89.3 & 87.5 & 85.1 & 86.4 & 84.6 & 86.8 & 92.4 & 65.5 & 62.2 & 82.8 \\    \cline{2-14}
 & \multirow{2}{*}{S$^{2}$I-TI-DIM} & w/o & 81.8 & 81.8 & 78.5 & 78.6 & 79.9 & 79.1 & 80.4 & 85.0 & 63.0 & 60.7 & 76.9 \\
 &  & w/ & 88.7 & 88.7 & 85.7 & 85.0 & 86.2 & 85.2 & 87.1 & 91.0 & 70.5 & 66.3 & 83.4 \\
 \hline
\multirow{12}{*}{Res-152} & \multirow{2}{*}{TI-DIM} & w/o & 55.7 & 55.9 & 52.5 & 43.4 & 55.3 & 46.5 & 51.9 & 66.0 & 46.5 & 31.4 & 50.5 \\
 &  & w/ & 58.7 & 58.7 & 55.3 & 43.6 & 56.4 & 47.2 & 55.6 & 70.7 & 48.4 & 33.0 & 52.8 \\ \cline{2-14}
 & \multirow{2}{*}{PI-TI-DI-FGSM} & w/o & 53.1 & 52.7 & 55.4 & 46.8 & 42.2 & 46.8 & 51.0 & 54.3 & 71.2 & 47.2 & 52.1 \\
 &  & w/ & 56.0 & 57.2 & 59.1 & 49.1 & 47.4 & 49.4 & 51.6 & 58.5 & 73.0 & 46.9 & 54.8 \\ \cline{2-14}
 & \multirow{2}{*}{VT-TI-DIM} & w/o & 67.9 & 73.4 & 69.3 & 60.5 & 70.4 & 64.9 & 68.3 & 80.7 & 52.7 & 43.0 & 65.1 \\
 &  & w/ & 74.2 & 77.1 & 72.6 & 64.8 & 74.8 & 67.7 & 71.8 & 83.2 & 55.6 & 46.8 & 68.9 \\ \cline{2-14}
 & \multirow{2}{*}{Admix-TI-DIM} & w/o & 79.0 & 83.0 & 78.7 & 70.4 & 80.2 & 74.8 & 78.3 & 86.4 & 65.2 & 57.7 & 75.4 \\
 &  & w/ & 81.1 & 86.1 & 82.1 & 73.2 & 81.4 & 77.5 & 81.0 & 88.4 & 67.4 & 59.3 & 77.8 \\ \cline{2-14}
 & \multirow{2}{*}{S$^{2}$I-TI-DIM} & w/o & 86.8 & 86.6 & 83.9 & 79.0 & 85.1 & 80.9 & 85.3 & 89.3 & 72.6 & 68.6 & 81.8 \\
 &  & w/ & 88.8 & 87.8 & 85.2 & 79.9 & 85.9 & 83.1 & 85.8 & 91.3 & 75.0 & 69.5 & 83.2 \\
 \hline\hline
\end{tabular}
}
\end{table*}


\textbf{Parameter:} In all experiments, the maximum perturbation is $\epsilon=16$, the iteration is $T=10$ and the step size is $\alpha=\epsilon/T$. For MI-FGSM~\cite{dong2018boosting}, we set the decay factor to $\mu=1.0$. For DI-FGSM~\cite{xie2019improving}, we set the transformation probability to $p=0.5$. For TI-FGSM~\cite{dong2019evading}, we set the kernel length to $k = 7$. For VT-FGSM~\cite{wang2021enhancing}, we set the hyper-parameter to $\beta
= 1.5$, and the number of sampling examples is 20.  For SI-NI-FGSM~\cite{lin2020nesterov}, we set the number of copies to $m_1 = 5$. For Admix~\cite{wang2021admix}, we set number of copies to $m_1 = 5$, the sample
number to $m_2 = 3$, and the admix ratio $\eta = 0.2$. For PI-TI-DI-FGSM~\cite{gao2020patch}, we set the amplification factor to $\beta
= 10$, the project factor to $\gamma = 16$ and the kernel to length $k_w=7$. For S$^{2}$-I-FGSM~\cite{long2022frequency}, we set the tuning factor for $\mathcal{M}$ as $\rho = 0.5$, and the number of spectrum transformations as $N = 20$. For our proposed SVD-based attacks, the Top-$k$ decomposed feature is $k=1$ and the hyperparameter is $\beta = 0.5$. The middle layers for the SVD-based feature decomposition are Mixed-6e for Inception-v3 (Incv3), the last Inception-B blocks of Mixed-6b for Inception-v4 (Inc-v4), the Mixed-6a for Inception-Resnet-v2 (IncRes-v2), and the last layer of block3 in Layer4 for Resnet-v2-152 (Res-152), respectively. 



\subsection{Attack Normally Trained Models}
In this section, we verify the effectiveness of our method against the normally trained models by training the adversarial samples with five different baselines. Based on the results in Table~\ref{tab:Normally-Trained}, we can make the following findings: 
\textbf{1)} For each source model, there are significant improvements in performance when comparing the results of different baselines with SVD to baselines without SVD. For example, when Inc-v3 is used as the source model, average attack success rates increase by 3.7\% (MI-FGSM), 3.5\% (DI-FGSM), 6.3\% (S$^2$I-FGSM), 3.0\% (SI-NI-FGSM), and 5.9\% (VT-MI-FGSM). 
\textbf{2)} When attacking black-box models, the adversarial transferability is also significantly improved after applying SVD. For instance, when VT-MI-FGSM and IncRes-v2 are used as the baseline and the substitute model, attack success rates increase from 81.1\% to 88.0\%, 76.6\% to 84.9\%, 66.7\% to 77.1\%, and 70.1\% to 79.2\% for transferring attack Inc-v3, Inc-v4, Res-152, and Res-50, respectively.



These findings can validate the effectiveness of our proposed method in improving attack performance against normally trained models. Moreover, when combined with existing attacks, it can further enhance the black-box transferability of adversarial examples.


\subsection{Attack Defense Models}
To further evaluate the effectiveness of our method, we evaluate the attack success rates against ten different defense strategies, including four adversarially trained models (Inc-v3$_{adv}$, Inc-v3$_{ens3}$, Inc-v3$_{ens4}$, IncRes-v2$_{ens}$) and six more advanced methods (HGD, R\&P, NIPS-r3, RS, JPEG, NRP). As shown in Table~\ref{tab:defense}, we can find that the proposed method significantly improves performance across source models with diverse training baselines. For example, the average attack success rates of Inc-v3 with TI-DIM, PI-TI-DI-FGSM, VT-TI-DIM, Admix-TI-DIM, and S$^2$I-TI-DIM are increased by 3.6\%, 2.6\%, 3.5\%, 3.4\%, and 3.4\%, respectively. The results indicate that the Top-1 decomposed feature contributes significantly to the final prediction. However, current defense strategies do not leverage these features to enhance robustness.

\subsection{Analysis of computational cost}
We calculate the additional computation cost for different baselines using Inc-v3 as the source model when attacking the 1000 images from the ImageNet-Compatible dataset. The results are presented in Table~\ref{tab:cost}. From the table, we can observe that our proposed SVD method introduces an additional computation cost of approximately 25-30\% for different baselines.

\begin{table*}[t]
\caption{The training cost for different baselines of attacking 1000 images from ImageNet-Compatible dataset with and without the proposed SVD-based attack}
\label{tab:cost}
\resizebox{\linewidth}{!}{
\begin{tabular}{c|ccccc|cccccc}
\hline
SVD & MI-FGSM & DI-FGSM & S2I-FGSM & SI-NI-FGSM & VT-MI-FGSM & TI-DIM & PI-TI-DI-FGSM & SI-NI-TI-DIM & VT-TI-DIM & Admix-TI-DIM & S$^{2}$I-TI-DIM \\
w/o & 23.39s & 27.60s & 723.84s & 110.09s & 452.02s & 27.83s & 40.58s & 125.96s & 525.35s & 348.98s & 815.24s \\
w/ & 30.02s & 34.98s & 890.65s & 143.37s & 594.03s & 35.16s & 50.23s & 164.40s & 684.05s & 438.35s & 998.17s \\
\hline
\end{tabular}}
\end{table*}

\subsection{Image Quality of Adversarial Examples}
We further compute the SSIM~\cite{wang2004image} scores between the original samples and the adversarial samples generated using the Inc-v3 model under different baselines, both with and without our proposed method. SSIM is a popular metric for evaluating image quality, a large SSIM score indicates better quality and less distortion.
As shown in Table~\ref{tab:ssim}, we can find that SSIM scores obtained without and with our proposed method are very similar. This suggests that our proposed method does not significantly alter the perceptual similarity between the original and adversarial samples.
\begin{table*}[t]
\caption{The SSIM between the original samples and the adversarial samples trained by different baselines without and with our proposed method.}
\label{tab:ssim}
\resizebox{\linewidth}{!}{
\begin{tabular}{c|ccccc|cccccc}
\hline
SVD & MI-FGSM & DI-FGSM & S2I-FGSM & SI-NI-FGSM & VT-MI-FGSM & TI-DIM & PI-TI-DI-FGSM & SI-NI-TI-DIM & VT-TI-DIM & Admix-TI-DIM & S$^{2}$I-TI-DIM \\
w/o & 0.8336 & 0.9505 & 0.9407 & 0.8313 & 0.8389 & 0.8112 & 0.7410 & 0.8110 & 0.8260 & 0.8079 & 0.8043 \\
w/ & 0.8332 & 0.9500 & 0.9425 & 0.8294 & 0.8617 & 0.8109 & 0.7414 & 0.8093 & 0.8444 & 0.8068 & 0.8034 \\
\hline
\end{tabular}}
\end{table*}
\subsection{Ablation Study}

\begin{figure}[h]
    \centering
    \begin{subfigure}{0.23\textwidth}
        \includegraphics[width=\linewidth]{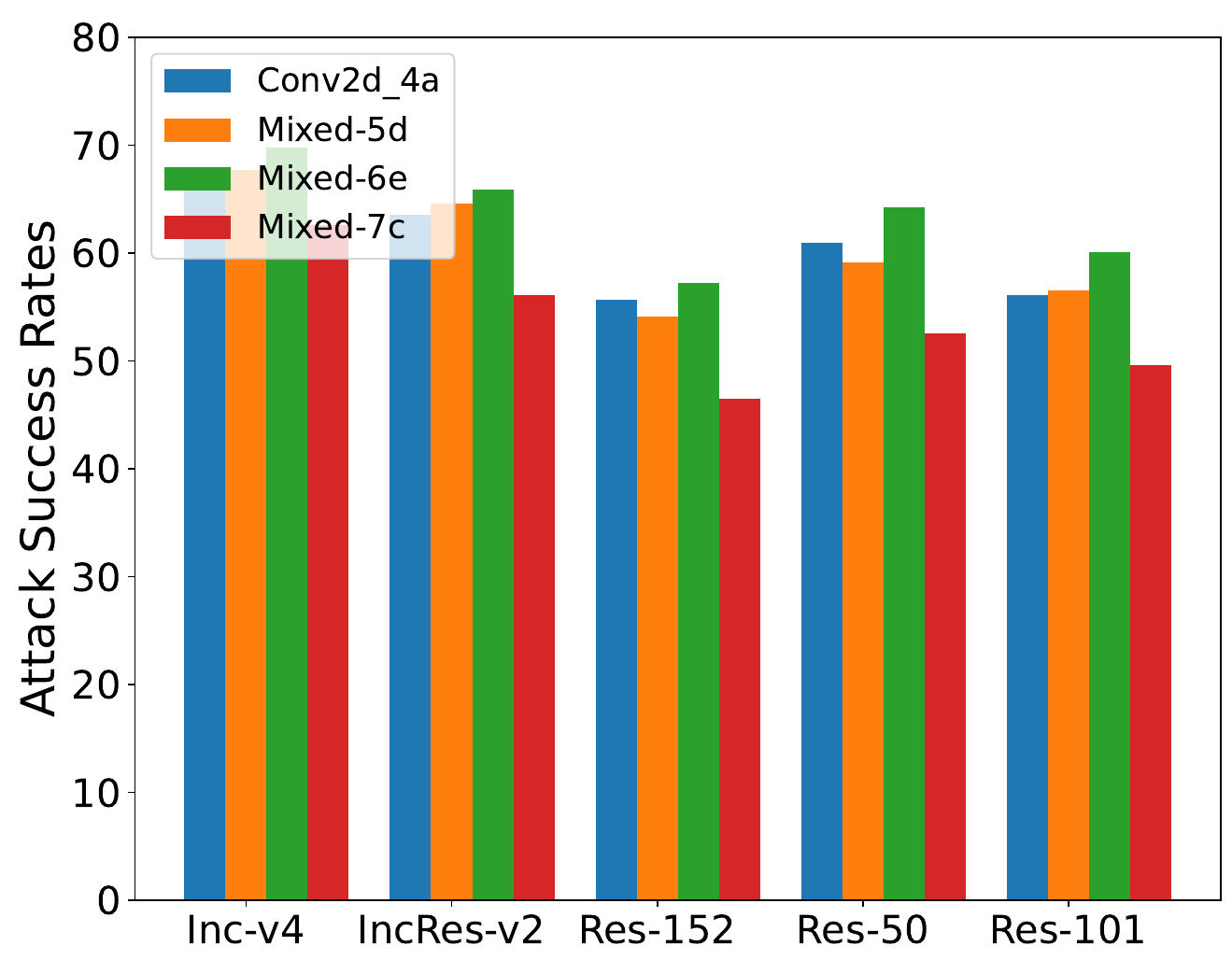}
        \caption{Inception-v3}
    \end{subfigure}
    \begin{subfigure}{0.235\textwidth}
        \includegraphics[width=\linewidth]{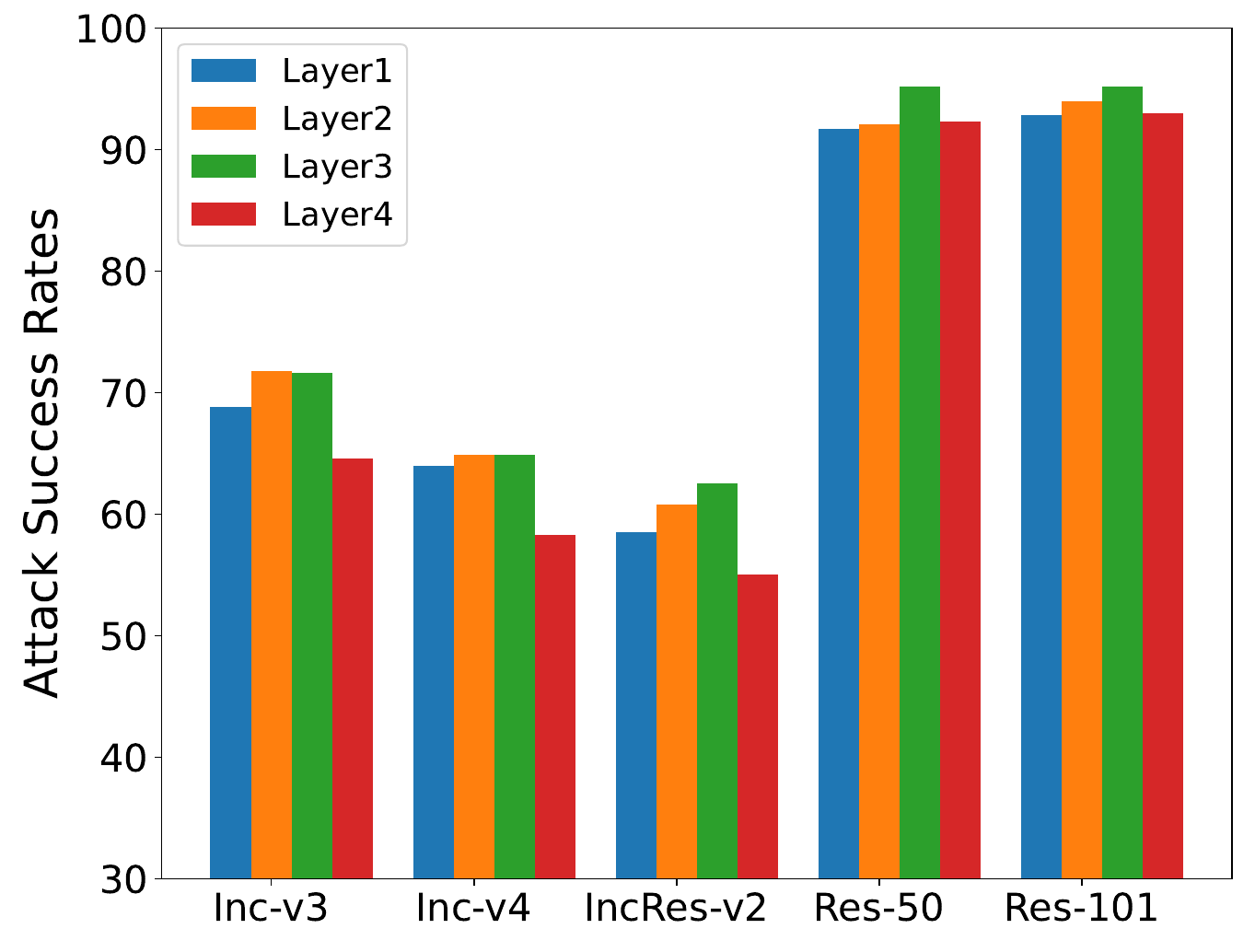}
        \caption{ResNet-152}
    \end{subfigure}
    \caption{The black-box performance of using different layers for feature decomposition.}
    \label{fig:layers}
\end{figure}

\textbf{Target Layer for Decomposition:} Firstly, we evaluate the performance using different layers for SVD feature decomposition. In Fig.~\ref{fig:layers} (a) and (b), we plot the transfer attack success rate using Inc-v3 and Res-152 as the source models, respectively. For each source  model, we compare four layers at four convolutional blocks with different feature resolutions. As shown in Fig.~\ref{fig:layers}, we can find that the best performance is achieved at Mixed-6e for Inception-v3 and Layer 3 for ResNet-152. The main reason for this phenomenon is mainly due to that the first two layers have high-resolution and less semantic information. The Mixed-7c and Layer-4 are the last convolutional layers in the Inc-v3 and Res-152 and have less generalization ability. And the decomposed feature with the largest singular value for the last layer is relatively close to the original features, which weakens the contribution of the logits combination.

\begin{figure}[h]
    \centering
    \begin{subfigure}{0.235\textwidth}
        \includegraphics[width=\linewidth]{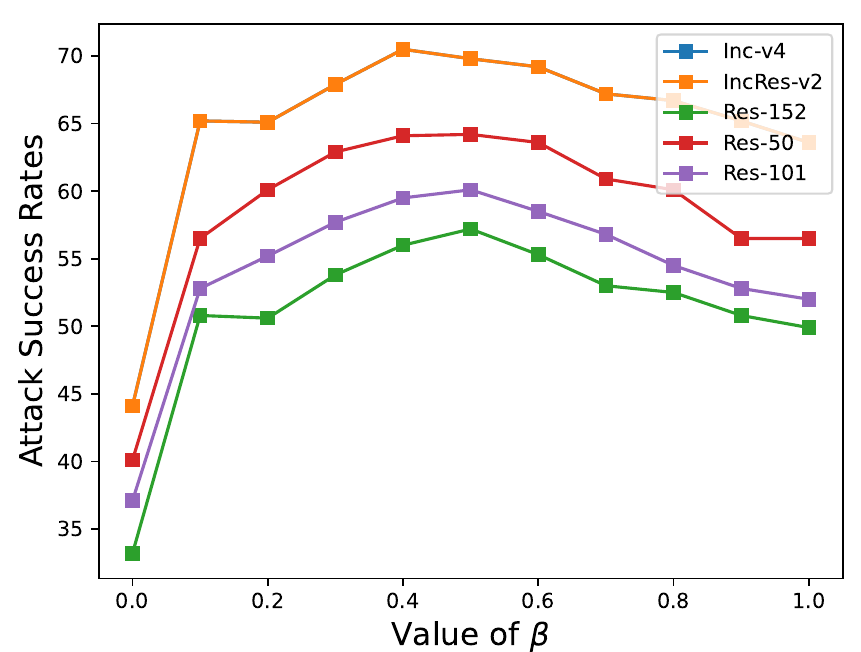}
        \caption{Sensitivity of Parameter $\beta$ }
        \label{fig:beta}
    \end{subfigure}
    \begin{subfigure}{0.235\textwidth}
        \includegraphics[width=\linewidth]{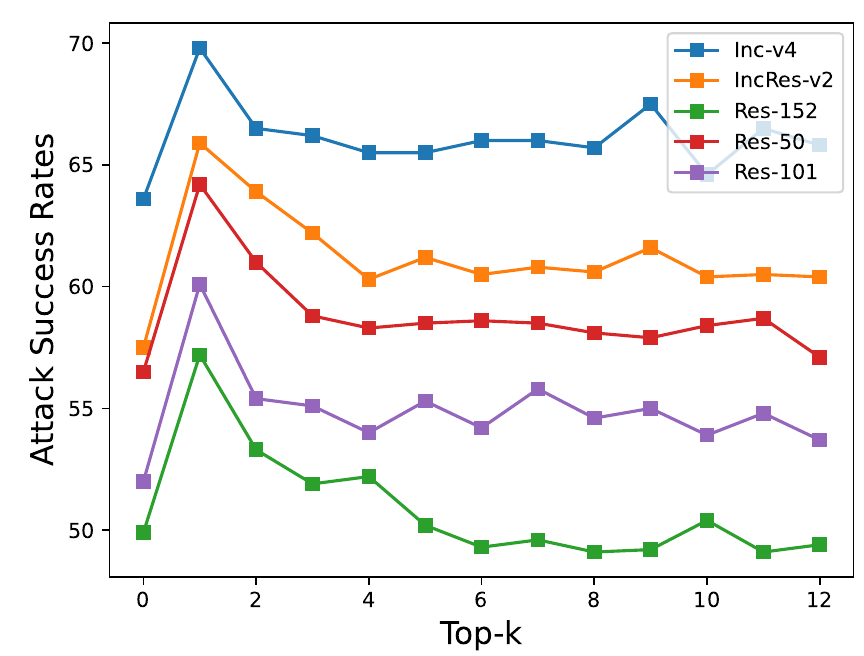}
        \caption{Influence of Top-$k$}
        \label{fig:top-k}
    \end{subfigure}
    \caption{Ablation studies of different hyper-parameters against Inception-v3 in the black-box setting.}
    \label{fig:logits_cal}
\end{figure}

\textbf{Parameter:} The trade-off influence of the hyperparameter $\beta$ for the Inc-v3 as source model is shown in Fig.~\ref{fig:beta}. We can find that the attack success rates increase along with the increment of $\beta$. The overall best performance is achieved at around $\beta=0.5$ for all five black-box models. When $\beta$ is increased further, the attack success rates decrease, indicating that using only the Top-1 decomposed feature is insufficient to achieve better performance. Besides, when $\beta=0$, original features are not utilized, resulting in significantly lower performance than other values. At $\beta=1$, SVD-based features are not utilized. Therefore, these results suggest the original feature is crucial for achieving better-attacking performance.

\textbf{Top-$k$:} In Fig.~\ref{fig:top-k}, we depict the performance of transferability under Top-$k$ decomposed feature $\bm{Z}_l = \sum_{i=1}^k s_i \bm{u}_i \bm{v}_i^T$. Compared with the original feature (\textit{i.e.}, Top-$0$), we can find that using the Top-1 decomposed feature can significantly boost the attack success rates in all models. However, performance will start to decrease when using Top-$k$ $(k>2)$ decomposed features. Besides, the attack success rates fluctuate around a certain value when $k>=5$, since $\sum_{i=1}^k s_i \bm{u}_i \bm{v}_i^T$ gradually equals to the original feature.

\subsection{Comparison with feature-based methods}
In this part, we compare our method with other feature-based methods in terms of both attack success rates and training costs. We craft adversarial examples using the same settings and data augmentation as NAA~\cite{zhang2022improving}.
Based on Table.~\ref{tab:compare-feature}, our method can achieve superior performance in terms of attack success rate compared to the FDA~\cite{ganeshan2019fda}, FIA~\cite{wang2021feature}, and NAA~\cite{zhang2022improving} methods. 
Besides, we compare the total training cost of using Inc-v3 as the source model on a 2080Ti GPU across 1000 images with other feature-based methods. As shown in Table~\ref{tab:cost}, our method is more efficient than the FIA~\cite{wang2021feature}, NAA~\cite{zhang2022improving}, and FDA~\cite{ganeshan2019fda}, which only requires 1029s for training.

\begin{table*}[h]
\caption{Comparing our method with feature-based SOTA methods. The attack success rates (\%) on four undefended models and four adversarially trained models.}
\label{tab:compare-feature}
\begin{tabular}{c|c|cccccccc}
\hline \hline
Model & Attack & Inc-v3 & Inc-v4 & IncRes-v2 & Res-152 & Inc-v3$_{adv}$  & Inc-v3$_{ens3}$ & Inc-v3$_{ens4}$ & IncRes-v2$_{ens}$  \\
\hline
\multirow{4}{*}{Inc-v3} & FDA & 81.3 & 42.9 & 36.0 & 35.4 & 19.3 & 8.9 & 6.4 & 2.3 \\
 & FIA & 98.3 & 83.2 & 79.1 & 71.6 & 53.3 & 36.1 & 37.0 & 20.0 \\
 & NAA & 98.1 & 85.0 & 82.4 & 77.1 & 61.5 & 50.5 & 50.8 & 31.5 \\
 & Ours & \textbf{99.8} & \textbf{87.8} & \textbf{85.2} & \textbf{80.7} & \textbf{64.2} & \textbf{61.2} & \textbf{58.8} & \textbf{38.6} \\
 \hline
\multirow{4}{*}{Inc-v4} & FDA & 84.8 & 99.6 & 71.9 & 68.7 & 27.9 & 18.4 & 17.2 & 7.3 \\
 & FIA & 84.1 & 95.7 & 78.6 & 72.0 & 45.3 & 38.0 & 37.2 & 19.4 \\
 & NAA & 86.0 & 96.5 & 81.0 & 75.5 & 52.4 & 50.5 & 49.4 & 30.8 \\
 & Ours & \textbf{90.5} & \textbf{99.5} &\textbf{ 86.3} & \textbf{80.9} & \textbf{60.8} & \textbf{65.7} & \textbf{63.4} & \textbf{48.1} \\
 \hline
\multirow{4}{*}{IncRes-v2} & FDA & 69.3 & 67.7 & 78.3 & 56.3 & 36.4 & 16.2 & 22.3 & 17.9 \\
 & FIA & 81.6 & 77.1 & 88.7 & 71.0 & 63.8 & 49.8 & 46.6 & 34.1 \\
 & NAA & 82.4 & 78.0 & 93.0 & 74.4 & 64.9 & 60.0 & 56.7 & 47.5 \\
 & Ours & \textbf{92.9} & \textbf{92.2} & \textbf{99.3} & \textbf{88.0} & \textbf{76.7} & \textbf{78.0} & \textbf{74.7} & \textbf{70.4} \\
 \hline
\multirow{4}{*}{Res-152} & FDA & 83.9 & 84.1 & 73.9 & 89.1 & 51.2 & 27.9 & 23.6 & 11.5 \\
 & FIA & 83 & 81.6 & 78.4 & 98.9 & 58.2 & 49.1 & 44.9 & 29.3 \\
 & NAA & 85.9 & \textbf{85.9} & \textbf{83.6} & 98.2 & 66.1 & 61.6 & 59.2 & 46.7 \\
 & Ours & \textbf{86.4} &84.8 &83.1 &\textbf{99.9}  & \textbf{70.7} &\textbf{ 70.4} &\textbf{ 67.6} & \textbf{57.2}\\
 \hline \hline
\end{tabular}
\end{table*}

\begin{table}[ht!]
\caption{The training cost comparison of our method with other feature-based methods.}
\label{tab:cost}
\begin{tabular}{c|cccc}
\hline
Method      & FDA~\cite{ganeshan2019fda}   &FIA~\cite{wang2021feature}    &NAA~\cite{zhang2022improving}    &Ours    \\ \hline
Cost Time (s)   & 1421 &7081 & 7163  &1029\\
\hline
\end{tabular}
\end{table}

\section{Conclusion}
In this paper, we propose the SVD feature decomposition-based attack method. Incorporating  decomposed features related to the largest singular value into the attack can intrinsically leverage the advantages of using more generalizable features and attentively localize the foreground object. Extensive experiments demonstrate that the proposed attack method can significantly boost the transferability for attacking both black-box models and defense strategies. On the other aspects, we could consider maintaining the clean singular vectors corresponding to the largest singular value for building more robust defense models.

\bibliographystyle{ACM-Reference-Format}
\bibliography{sample-base}

\end{document}